# VQUNet: Vector Quantization U-Net for Defending Adversarial Attacks by Regularizing Unwanted Noise

Zhixun He
University of California, Merced
Merced, CA, U.S.A
zhe5@ucmerced.edu

Mukesh Singhal
University of California, Merced
Merced, CA, U.S.A
msinghal@ucmerced.edu

## ABSTRACT

Deep Neural Networks (DNN) have become a promising paradigm when developing Artificial Intelligence (AI) and Machine Learning (ML) applications. However, DNN applications are vulnerable to fake data that are crafted with adversarial attack algorithms. Under adversarial attacks, the prediction accuracy of DNN applications suffers, making them unreliable. In order to defend against adversarial attacks, we introduce a novel noise-reduction procedure, Vector Quantization U-Net (VQUNet), to reduce adversarial noise and reconstruct data with high fidelity. VQUNet features a discrete latent representation learning through a multi-scale hierarchical structure for both noise reduction and data reconstruction. The empirical experiments show that the proposed VQUNet provides better robustness to the target DNN models, and it outperforms other state-of-the-art noise-reduction-based defense methods under various adversarial attacks for both Fashion-MNIST and CIFAR10 datasets. When there is no adversarial attack, the defense method has less than 1% accuracy degradation for both datasets.

## 1 INTRODUCTION

Modern DNNs have made their success in achieving excellent performance on a wide range of tasks in various domains, e.g., images [14, 18, 27], audio [22, 43] and videos [7, 13], etc. However, it's been widely observed that DNN-based approaches are sensitive to samples perturbed by adversarial attacks [1, 6]. Such adversarial samples can be indistinguishable from real data but they can cause DNN models to give wrong predictions. Because each benign sample can be used to craft various adversarial counterparts, it's challenging for a defense mechanism to be able to handle all possible adversarial samples without introducing any decrease in a model's prediction performance on benign data when there is no adversarial attack.

Adversarial Training is one defense approach for DNN models to mitigate the attacks, by mixing adversarial samples with benign data [4, 11] when training a DNN. Although it has shown progress in robustness improvement, there are two main drawbacks: (1) each real sample can be crafted by various attack algorithms and it is difficult to include all possible adversarial samples into training samples; (2) there is a trade-off between the diversity of adversarial samples and training time to achieve satisfactory generalization to tackle all adversarial attacks. Another line of effort is to certify a robustness threshold for DNN models [37], that given an adversarial sample, as long as the difference of itself from its benign counterpart is within a $L_p$-norm-ball, the certified model will be resistant to the adversarial sample no matter what attack algorithm is used to create it. In practice, in order to simulate various adversarial attacks and the adversarial noise that covers the $L_p$-norm-ball on benign samples, a random sampling process is used to create adversarial samples during training.

Other than adversarial training, various other categories also show progress in adversarial defense. (1) Gradient Obfuscation, which refers to mechanisms that inhibit the adversary to calculate the adversarial noise. For example, Papernot et al. [36] dynamically adjusts the activation function to reduce the magnitude of the gradient of the loss with respect to the model's input such that the adversarial perturbation is reduced. (2) Detection, in which the benign training data can maintain their consistency in their latent space features, and this consistency can distinguish themselves from the latent feature anomaly hidden in adversarial counterparts [8, 48]. (3) Adversarial Noise Reduction, which takes unknown inputs to reconstruct outputs that are similar to the original benign data by reducing unwanted adversarial noise in the unknown inputs. For example, generative neural networks, like, VAE [24, 25] and GAN [10], are common paradigms for generating noise-reduced samples by mapping inputs to the outputs from the desired data domain.

VAE-based noise reduction defense methods [9, 15, 53] optimize their performance by reducing the divergence between reconstructed output and the training data. Although this objective allows VAE to generalize well to unseen data, VAEs with sequential structure typically suffer from unwanted noise during the decoding stage, e.g., incurring blur effect in the reconstructed images. GANs [2, 23, 45] demonstrated promising generative features, which can create quality images with high resolution. While both generative networks hold the promise of generating noise-reduced images for defense purposes, it's still a challenge to reconstruct samples as clean as original training data when the adversarial noise becomes more noticeable. Also, the more complicated the dataset the more difficult for the generative networks to maintain the data reconstruction quality (e.g., RGB images from Cifar10 versus gray-scaled images from Fashion-MNIST), thus, the performance of noise-reduction-based methods vary for different datasets [15, 16, 20, 21].

In this paper, we present a new generative model for adversarial noise reduction that combines multi-scale hierarchical representation learning and discrete latent feature learning through Vector Quantization (VQ), to reconstruct high-fidelity images and achieve high tolerance to adversarial noise. In particular, this work makes the following contributions:

- We introduce Vector Quantization U-Net (VQUNet), a novel generative model that maintains the data fidelity after noise reduction, by leveraging multi-scale hierarchical latent feature learning and quantized feature learning. VQUNet further improves the robustness of data reconstruction under various adversarial attacks and it outperforms existing state-of-the-art adversarial defense methods, especially when the adversarial perturbation is high.
- Via rigorous experiments, we show that VQUNet outperforms state-of-the-art adversarial noise reduction defense methods across a wide spectrum of adversarial noise levels and various adversarial attacks on two benchmark datasets (CIFAR10 [26] and Fashion-MNIST [46]). The de-noising process through VQUNet does not sacrifice the prediction accuracy on clean data after they are processed by the de-noising network. The proposed method incurs a minimal impact on the original deep learning model's performance when there is no adversarial attack.
- Through extensive experimental study, we further examine the individual components and provide insight into their effectiveness on the overall performance against adversarial attacks.

## 2 RELATED WORK
### 2.1 Adversarial Noise Reduction

Adversarial noise-reduction-based methods aim to clean the input before they are predicted by a deep learning model [47]. The cleaning process significantly reduces the effectiveness of adversarial attacks on the target models [1]. For example, during the JPEG compression/decompression process [30], by quantizing the images' frequency domain, a certain level of adversarial perturbation can be filtered. Other than reducing noise for the entire image, adversarial noise reduction can also be applied only to a high-attention region in an image using a wavelet-based image de-noising process [12].

Sun et al. [42] tackle adversarial noise by projecting an input to a feature space using a "Sparse Transformation Layer", and approximating the projected space to that of natural images. In comparison, Defense-GAN [41] tries to find a latent space feature that can be used by the generator to reconstruct the image as close as the testing input, and then this latent feature will be used as the seed to generate the de-noised version of the input.

Yuan et al. [49] mitigate the adversarial noise pattern by processing an input image through an orthonormal transformation and a deadzone-based activation function. Then a generative network with a feedback loop provides multiple estimations that will be used in a dedicated neural network to restore the clean image. He et al. [15] improve the prediction accuracy on adversarial samples by combining the prediction accuracy on noise-reduced images and a majority voting on multiple estimations through a Bayesian update process.

Multiple VAE-based noise reduction methods [9, 19, 29] attempt to use VAEs multiple times to further reduce the adversarial noise. However, there is a caveat to reusing the same de-noising network multiple times. If the first pass introduces an unwanted effect in the output, e.g., a blur effect, such effect can be amplified when passing the output into the de-noising network for a second time. A different attempt enforces a cycle consistency loss when training two de-noising generative networks simultaneously [16]. In this process, one generative network generates images to be in a different domain from the original domain, and then a second generative network restores the generated outputs back to the original images' domain.

### 2.2 Generative Networks on Adversarial Defense

Although adversarial noise-reduction-based methods can mitigate adversarial attacks, special care should be given to the data reconstruction process, because it is important for the DNN model to keep the same prediction accuracy for benign inputs after the de-noising process. Maintaining the fidelity of the reconstructed data while effectively reducing the adversarial noise is still a challenge.

VAE [24, 25, 51] and GAN [10, 34, 45] are two of the most popular generative network paradigms for generating samples. VAE features a two-step process to reconstruct the data. The inputs are first encoded into latent feature space, and then the encoded vectors are decoded back to the original data space [52]. This encoding/decoding feature is attractive for defense purposes as the encoding process can suppress the adversarial noise [9, 15, 29, 32, 53]. However, the decoder may struggle to recreate high-fidelity data if the encoding process discards too much information, and it can introduce unwanted noise in the reconstructed images.

As opposed to data encoding/decoding, GANs aim at generating outputs that are similar to those from the original domain, by training a generator and a discriminator simultaneously. The discriminator attempts to distinguish the actual training samples from the generated samples, and the generator tries to generate outputs that can confuse the discriminator [33]. The relaxation from the data encoding/decoding gives GANs flexibility and diversity in their network structure and training schemes [2, 16, 20, 21, 23, 34]. For example, U-net [40] designs their feature extraction and data reconstruction path to be symmetric at different depth levels as opposed to traditional linear expansion.

### 2.3 Quantized Vector as Latent Features

Multiple studies [3, 19, 30, 48] have indicated that data compression has a regularizing effect of reducing model sensitivity for small noise to mitigate adversarial attacks. Input discretization is one of the compression techniques, which refers to the process of converting continuous values into different non-overlapping buckets [3]. Different discretization strategies have been studied, such as, evenly quantizing the compression feature [30], adjusting the quantization process dynamically [38], and discretizeing the feature vector into a vector dictionary [42, 44]. The proposed method learns a codebook that consists of multiple quantized vectors, and each of the original feature vectors will be mapped to one of the vectors in the codebook based on the similarity of the original vector and the vectors in the codebook.

The input quantization process has a regularization effect on adversarial noise, which is partially due to the fact that a certain level of unwanted adversarial noise will be discarded when compressing input to a smaller dimension. However, there is a trade-off between the compression rate and the data reconstruction quality. If too much information is discarded, the reconstruction process will struggle to generate samples with high fidelity to the inputs, especially when the dataset becomes more complicated, for example, images from CIFAR10 are more challenging to be reconstructed as opposed to those from Fashion-MNIST or MNIST.

The vector quantization model [44] features a learned codebook that contains up to $K$ 1-D vectors. Toward the end of the encoding process, each of the natural feature vectors will be mapped to one of the vectors in the codebook. The non-linear mapping is decided by: 1) calculating the Euclidean distance between the natural feature vector and the quantized vectors in the codebook; 2) the quantized vector with the closest distance will be selected as the replacement for the natural feature vector. Formally [39], with the input data, $x$, to quantize the natural feature vectors $E(x)_i$, the quantized feature $Q(E(x)_i)$ is calculated as below:

$$Q(E(x)_i) = e_k \quad \text{where} \quad k = \arg\min_j \left\| E(x)_i - e_j \right\|_2 \quad (1)$$

where $k \in 1, 2, \cdots, K$ (K is the total number of quantized vectors in the codebook), and $i \in 1, 2, \cdots, M$ where $M$ is the total number of natural feature vectors in $E(x)$. The quantized feature vectors $Q(E(x))$ will be used for data reconstruction during the forward pass. The training of the codebook will be further explained in section 3.2.

## 3 METHOD

In this work, we introduce a new adversarial noise reduction network, Vector Quantization U-Net (VQUNet), an effective noise reduction model to regularize both adversarial and clean data. VQUNet enhances deep learning models' robustness against various adversarial attacks. VQUNet features a U-shaped latent feature forward pass and hierarchical vector quantization to capture the correlation between high-level latent features and low-level information, such as texture, localized features, and geometry of objects. The combination of the information from different hierarchies of the network allows the decoding blocks to propagate the local texture features toward the end of the forward pass to yield a more precise recreation of the input.

During the forward pass, VQUNet applies non-linear mapping from each encoding block's natural outputs to discrete codes at multiple hierarchical levels. The discrete codes are selected from a learnable codebook, where each natural feature vector is mapped to the most similar discrete code. In comparison, the classic VAEs [9, 24] compress latent code (the prior $p(z)$ under the Bayesian theory) into a continuous multi-dimensional unit ball, while the VQUNet converts each natural feature vector into one of the learned vectors in the codebook.

### 3.1 VQUNet Structure

The structure of VQUNet is illustrated in Figure 1a, where the input is an image, and the output is its noise-reduced version. For simplicity, we annotate the encoding, decoding, and quantization process with $E_d$, $D_d$, and $Q_d$ respectively, where $d$ represents their depth level. The single arrows with labels indicate the data flow in Figure 1a, and the details of the structure of $E_d$, $D_d$ are illustrated in Figure 1b and Figure 1c respectively.

The vertical direction in Figure 1a is the encoding process, which extracts more global shared latent features as the forward pass moves deeper vertically. A common practice for latent feature extraction is to apply a sequence of Convolutional Layers to the feature outputs. However, when the forward pass is too deep, the network struggles to use the highly abstract latent features for data reconstruction. Therefore, we depart from this design by applying residual blocks to the encoding process, which helps maintain the information from earlier layers, avoiding the information degradation problems, as shown in Figure 1b.

Each encoded feature vector from the encoder block will be mapped to one of the quantized vectors in the codebook, where the quantized vector with the closest Euclidean distance from the feature vector will be selected. Note that each depth level has its own quantization process and unique codebook for the quantization process. The new set of quantized vectors selected from the codebook at each depth level will be used as the input for the next level's encoding process.

At the deepest level after all encoding processes, the latent features have been transformed to the dimension of $2 \times 2 \times 512$. From there, the decoding process starts to expand the highly abstract information, which includes two stages: (1) the features first go through a convolutional block and two residual blocks; (2) then a ConvTranspose process expands the size with a stride of two; as shown in Figure 1c.

The decoded features will be concatenated to the quantized features from the immediate previous depth level. Then these hybrid features are expanded again by the next decoding block as the data pipeline shows in Figure 1a. Toward the end, the size of the output reaches that of the original data, and the output's channel number will be adjusted accordingly to be consistent with the domain of the original data.

### 3.2 Training

For the $N$ images $X = \left\{ x^{(i)} \right\}_{i=1}^{N}$, the VQUNet regularizes the input to its noise-reduced counterpart $X^{\sim}$: $f_{VQUNet} : X \rightarrow X^{\sim}$. The adversarial version of the data is denoted as $X^{adv} = \left\{ x^{adv(i)} \right\}_{i=1}^{N}$.

**Reconstruction Loss.** For the data reconstruction, the loss to guide the network to recreate data that are similar to the clean samples is the mean square error between the inputs and network outputs:

$$L_{reconst} = \mathbb{E}_{x \in X} \left[ \left| x - f_{VQUNet}(x) \right|^2 \right] \quad (2)$$

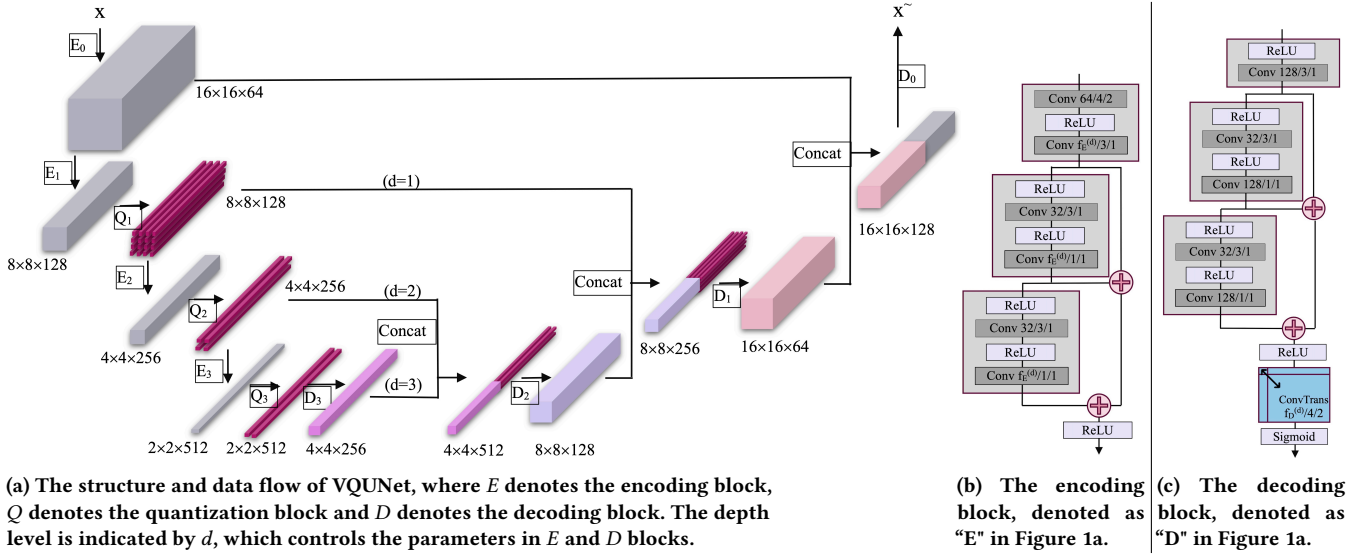

(a) The structure and data flow of VQUNet, where $E$ denotes the encoding block, $Q$ denotes the quantization block and $D$ denotes the decoding block. The depth level is indicated by $d$, which controls the parameters in $E$ and $D$ blocks.

(b) The encoding block, denoted as "E" in Figure 1a.

(c) The decoding block, denoted as "D" in Figure 1a.

Figure 1: The overall structure of VQUNet, and the detailed structure of encoder block and decoder block. The three-digit pair in Conv. and ConvTrans., such as (32/3/1), in Figure 1b and 1c represents (output filter number / kernel size / strides )

The encoding process is denoted as $f_E : x \rightarrow a$. $a_d$ is used to denote the output from an encoding block at depth level $d$. The vector quantization process is denoted as $f_Q : a \rightarrow q$, and the quantized vector at depth level $d$ is denoted as $q_d$. Note that the implementation of the codebook is through the Embedding Layer in Tensorflow, in which a 2-D matrix is created as the codebook parameters. The size of a 2-D matrix is $K \times G$, where $K$ is the total number of the 1-D quantized vector in the codebook, and each quantized vector has $G$ entries. The first step of the quantization process is to compare an input vector against all the $K$ discrete vectors, such that the index of the most similar discrete vector can be found. Then this index is used in Embedding Layer's look-up process to output the corresponding discrete vector.

However, this look-up operation that compares the input vector against $K$ discrete vectors and then finds the index of the most similar discrete vector, is a discrete process. Because of the discrete nature of the comparison operation, in the Tensorflow computation graph of the neural network, there is not a differentiable path during gradient back-propagation from the Embedding Layer to its previous layer. In such case, the gradient of $L_{reconst}$ w.r.t. the parameters of those layers before the Embedding Layer (e.g., the parameters in encoding blocks) will not be available, for example, Tensorflow assigns "None" or "Zero" to the gradient of loss w.r.t. those variables because they are regarded as unconnected variables.

To bypass the non-differentiable path issue above, an intermediate variable is used to bridge the encoding blocks' parameters to the Embedding Layer, such that the encoders' parameters are connected to the rest of the computation graph during the gradient back-propagation. First, $a_d$ and $q_d$ are already available during each forward pass, and a new variable $o_d$ is created in the computation graph and defined as $o_d = q_d - a_d$. The value of $o_d$ is treated as constant, and then a new quantized vector is created and defined as $q_d^* = a_d + o_d$. $q_d^*$ will be used as the input for the layer that comes immediately after the quantization block at depth $d$.

Note that $q_d^*$ holds the same value as $q_d$. However, by treating $o_d$ as constant and using $q_d^*$ in the forward-pass instead of $q_d$, the computational path from $a_d$ to $q_d^*$ becomes linear. This linear relationship makes the back-propagation path from $q_d^*$ to $a_d$ differentiable in the network. Thus, the parameters in Encoder blocks can be optimized with the gradient of $L_{reconst}$ during training.

Optimizing the quantized vectors in the codebook w.r.t. $L_{reconst}$ is done by normal back-propagation, because the quantized vectors, $q_d$, already form a continuous connection to the layers that come immediately after them.

**Encoder&Quantization Loss.** Because there is no constraint on the value range for $a_d$ and $q_d$, as training goes, $a_d$ and $q_d$ can end up having large differences in their elements' value magnitude.

This value range divergence can make it difficult for the training to converge. For example, if the values of the elements in $a_d$ are much bigger than that of $q_d$, then during the discrete mapping, a small change in $a_d$ can cause a very different set of quantized codes to be selected between each training step. Therefore, two extra loss functions are imposed to bring the value range of $a_d$ and $q_d$ closer.

First, to guide the parameters of the encoder to be closer to the codebook's value range, the intermediate quantized vectors $q_d^*$ will be reused. The mean square error between $q_d^*$ and $a_d$ is used as the loss function to guide the encoder's outputs to be closer to the values of the quantized vectors in the codebook. The encoder loss is defined as:

$$L_E = \mathbb{E}_{x \in X}\left[\left|q_d^* - a_d\right|^2\right] \tag{3}$$

Second, in order for the quantized vectors in the Embedding Layer to form a similar value range to the encoder's outputs, another intermediate variable, $a_d^*$, is created in the computation graph for training. The variable $a_d^*$ is used to hold the value of the encoder's outputs, $a_d$. Then the mean square error between $a_d^*$ and $q_d$ is used as the quantization loss to guide the training of the quantized vectors in the codebook, denoted as $L_Q$:

$$L_Q = \mathbb{E}_{x \in X} \left[ \left| q_d - a_d^* \right|^2 \right] \quad (4)$$

The overall loss to guide the training of VQUNet is as follows:

$$L = \alpha * L_{reconst} + \beta * L_E + L_Q \quad (5)$$

where $\alpha$ and $\beta$ are two weight parameters that can be adjusted to give different priority to the individual loss. During the empirical study, the value of $\alpha$ and $\beta$ affect the convergence speed of $L_{reconst}$, $L_E$, and $L_Q$ only around the first 30 epochs, but after that, the convergence of loss does not vary much for different values of $\alpha$ or $\beta$. Therefore, for the rest of the experiments, $\alpha = 1$ and $\beta = 1$ are used.

## 4 EXPERIMENTS

### 4.1 Experimental Setup

Once VQUNet is trained, the de-noised version of data, $X^\sim$, shall be used to retrain the target deep learning model. During the testing phase, VQUNet acts as an adversarial noise filter for the inputs, and then the de-noised outputs are fed into the target deep learning model for testing. In our experiments, two datasets are used for evaluation: CIFAR10 [26] and Fashion-MNIST [46]. Each sample in CIFAR10 is a 32 × 32 × 3 RGB-colored image, and each sample belongs to 1 of the 10 categories of objects. Individual sample in Fashion-MNIST is a 32×32×1 gray-scaled image, and there are also 10 categories of fashion objects in the dataset. CIFAR10 contains 50,000 training samples and 10,000 testing samples, and Fashion-MNIST contains 60,000 training samples and 10,000 testing samples.

For comparison, the proposed method and three other existing adversarial noise reduction methods are evaluated: the proposed VQUNet, Defense-VAE [29], High-Frequency Loss VAE (FHL_VAE) [15], and Defense-CycleGAN (CycleGAN) [16]. Following a common practice for adversarial defense evaluation, all defense methods will be tested under four different adversarial attacks: Fast Gradient Sign Method (FGSM) [11], Basic Iterative Method (BIM) [28], Projected Gradient Descent (PGD) [31], Carlini and Wagner Method (CW) [5]. All noise reduction networks will only be trained once and their performance is evaluated against all four adversarial attacks for a wide range of noise levels $\varepsilon$. Note that there is a post-training add-on feature that can be attached to most of the noise reduction process to further boost the target model's accuracy as mentioned in [15], but to test the data reconstruction properties, the experiments did not integrate the add-on feature but just focus on the noise reduction properties.

The implementation of generating adversarial samples for all four different attacking algorithms is through Cleverhans [35] package. White-box attack refers to the case when the adversary has full access to the target models' information, e.g., parameters, structure, and hyper-parameters, to generate adversarial samples. For a black-box attack, in comparison, the attacker does not have any knowledge about the target network, but the attacker can test the target model through trial and error, such as collecting the results using adversarial inputs computed from other networks. In the experiment, 50% of the testing samples are created using white-box attack algorithms on the target model, which is a Wide Residual Network (with parameters $depth$ = 28 and $k$ = 10), and the other 50% are created with 3 trained DenseNet [17] with different structures.

When creating adversarial samples using different attack algorithms, we keep most of the parameters to their default values, except the noise perturbation level $\varepsilon$. For classification accuracy, the target model being tested is a Wide-Residual-Network [50].

### 4.2 Accuracy Degradation from Noise Reduction Process

Note that the noise reduction defense methods filter every input before the de-noised output is used by the target model, even when there is no adversarial attack. We use "filtered" and "unfiltered" to distinguish the data that have been processed by the noise reduction networks as opposed to those data that have not. We call those data that have not been modified by an attack algorithm as "benign data", and those that are crafted by an attack algorithm as "adversarial samples".

It is important to see if the filtering process would change the target model's original performance, hence, we tested the target model's accuracy on the benign data before and after the filtering process, and the result is shown in Table 1.

Compared with the other three defense methods in Table 1, "VQUNet" is the only one that has < 1% accuracy degradation for both CIFAR10 and Fashion-MNIST datasets, which means "VQUNet" incurs the least alteration to the target model's original performance.

| model's acc on benign CIFAR10: 94.03% | | |
| --- | --- | --- |
| Methods | acc (filtered benign) | acc degradation |
| VQUNet | **93.08%** | **-0.95%** |
| Cycle_GAN | 91.78% | -2.25% |
| HFL_VAE | 86.51% | -7.52% |
| Defense_VAE | 78.84% | -15.19% |
| model's acc on benign Fashion-MNIST data: 94.31% | | |
| Methods | acc (filtered benign) | acc degradation |
| VQUNet | **93.39%** | **-0.92%** |
| Cycle_GAN | 93.13% | -1.18% |
| HFL_VAE | 92.81% | -1.50% |
| Defense_VAE | 92.83% | -1.48% |

Table 1: The target model's accuracy before and after the noise reduction process of different defense methods on CIFAR10 (top) and Fashion-MNIST (bottom).

### 4.3 Defense against White/Black-Box Attacks

The comparison of the four noise reduction defense methods against various adversarial attacks on CIFAR10 and Fashion-MNIST datasets is shown in Figure 2. Note that when there is no defense, the target model's accuracy drops below 10% under all adversarial attacks for both datasets.

**CIFAR10.** From the result shown in Figure 2 (top), "VQUNet" maintains significantly higher accuracy than the other defense methods across the entire adversarial noise levels. As adversarial noise perturbation increases, "VQUNet" shows a flatter curve in accuracy compared to other methods, which indicates that "VQUNet" has better robustness against adversarial noise.

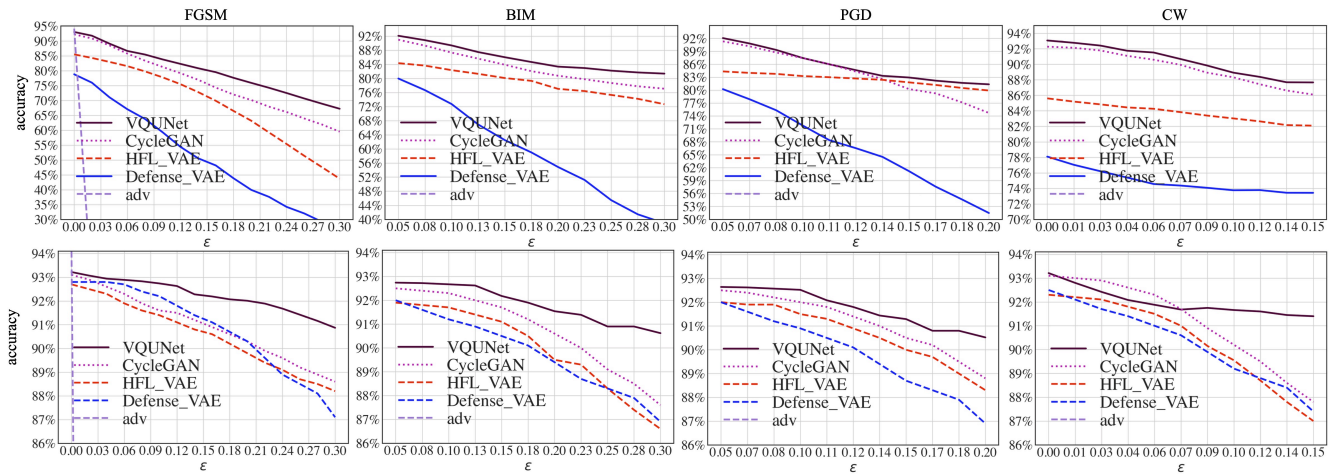

Figure 2: Comparison of different defense methods' performance under various adversarial attacks on CIFAR10 (top) and Fashion-MNIST (bottom)

**Fashion-MNIST.** Defense performance of different defense methods on Fashion-MNIST is shown in Figure 2 (bottom). We observe that compared to other methods, "VQUnet" demonstrates a flatter accuracy curve as the noise perturbation level increases. This indicates that "VQUnet" has a better resistance against adversarial noise for the Fashion-MNIST dataset. Among the four methods in our experiment, "VQUnet" is the only one that maintains at least 90% accuracy under all adversarial attacks across the entire noise spectrum.

### 4.4 Regularization from Vector Quantization

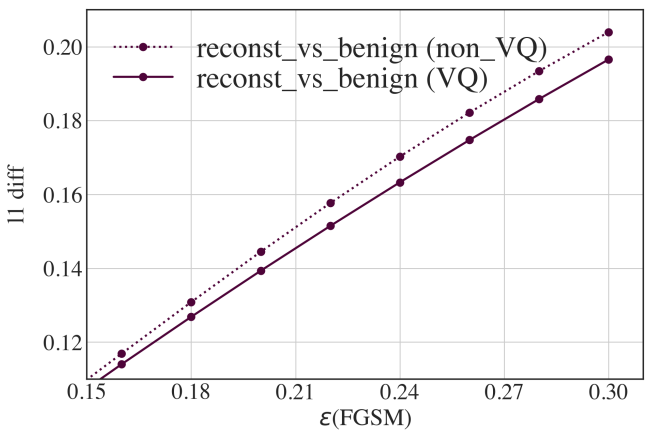

Figure 3: The $l1$ difference between benign images and the reconstructed images that are generated from VQUNet and non-VQUNet for various adversarial (FGSM) noise levels.

**VQ v.s. Non-VQ.** One of the main motivations for using vector quantization is to regularize the impact of adversarial noise on latent features. To study the regularization effect from VQ, we compared VQUNet to a second generative network called "non-VQUNet", which is nearly identical to the VQUnet except that there is no VQ block in it.

We first compared the averaged $l1$ difference between the original images and the reconstructed images from the generative networks when they were under FGSM attack. The results for both VQUNet and non-VQUNet are shown in Figure 3. As the adversarial perturbation increases, the reconstructed images from VQUNet demonstrated less divergence from the original images compared to those of the non-VQUNet.

To take a closer look, our experiments also examine the behavior of encoding blocks for both VQUnet and non-VQUNet when they are under FGSM attack. Figure 4 shows the $l1$ difference between the encoding blocks' output before and after the network is under FGSM attack at various depth levels (d=1, d=2, and d=3) for both of the VQUNet and non-VQUNet. The value deviation of the encoders' output before and after the attack for the non-VQUNet is much larger than that of VQUNet, which means the encoding features in the non-VQUNet are much more susceptible to adversarial noise than those in VQUNet. This helps explain why the VQUNet achieves better fidelity in image reconstruction as shown in Figure 3.

**Pre_VQ v.s. Post_VQ.** Note that during the VQ process, the pre_vq features are mapped to post_vq features through the VQ layer. In order to see the regularizing effect of VQ on the adversarial noise, we compared the $l1$ difference of the features before and after the adversarial attack (FGSM) for both pre_vq vectors and post_vq vectors at different depth levels, as shown in Figure 5. It's observed that both pre_vq and post_vq vectors show an increase in the divergence of their features before and after the attack as the adversarial noise level becomes larger. But for each depth level, the $l1$ difference in post_vq vectors is noticeably smaller than that of pre_vq. In other words, the adversarial noise has less negative impact on the post_vq vectors after the VQ process, compared to those of pre_vq vectors. Comparing different depth levels, both

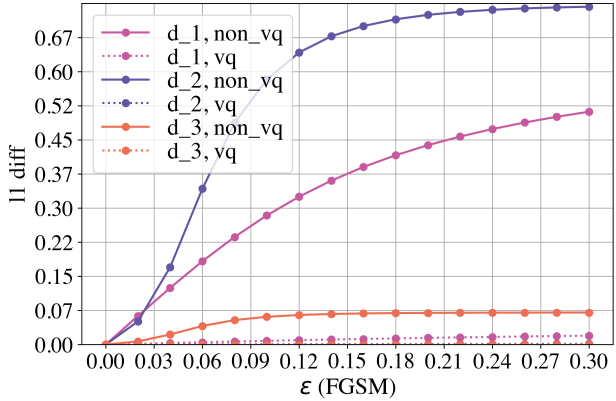

Figure 4: The $l1$ difference of pre_vq vectors before and after FGSM attack at various depth levels.

pre_vq and post_vq at deeper depth show less susceptibility to adversarial noise compared to those at shallower depth.

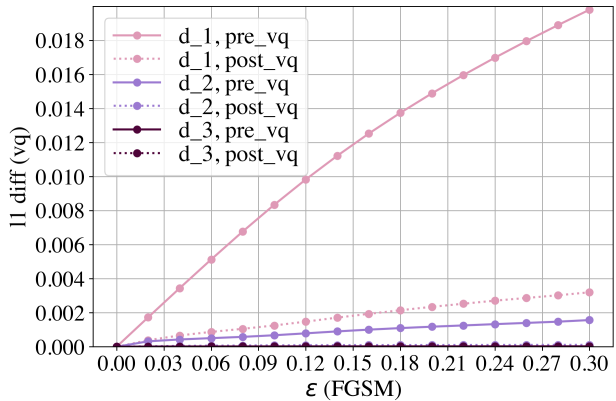

Figure 5: L1 difference of features before and after the FGSM attack for pre_vq and post_vq vectors at various depth levels.

**Code Index Change under Adversarial Attack.** Note that the VQ process maps the encoder blocks' outputs to a new set of discrete codes, where each code has its unique index in the codebook. To study the effect of adversarial noise on the discretization process, we compared the how much percentage change of the selected codes' indices before and after the adversarial attack (FGSM). Figure 6 shows the percentage change of selected codes' indices at different depth levels (d=1, d=2, and d=3) of VQUNet when it's under the FGSM attack. From the observation, the VQ layer at a deeper level maintains its original codes' indices relatively better than that at a shallower depth. As the adversarial noise increases, the change of indices rises more rapidly at the beginning, but this change slows down as the adversarial noise level becomes larger. This means that the VQ process shows more susceptibility to noise perturbation when the noise level is small. However as the adversarial level becomes larger, the VQ process maintains a similar selection of the discrete codes when it's under the attack.

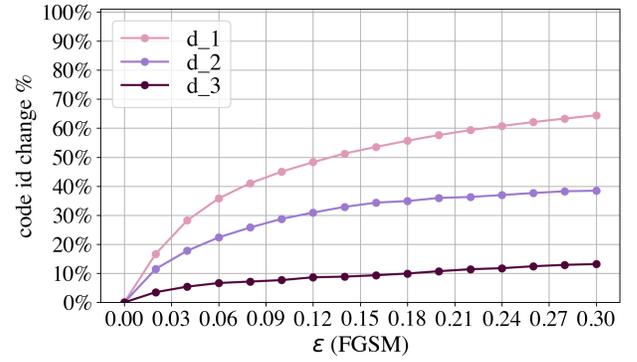

Figure 6: The percentage change of the indices of selected discrete codes before and after the FGSM attack.

## 5 CONCLUSION

We introduced a new noise-reduction model, VQUNet, for adversarial defense. The features include a hierarchical structure for high-fidelity image reconstruction and a vector quantization mechanism that helps regularize the impact from the adversarial noise for better data reconstruction. We showed that under various adversarial attacks and a wide range of noise perturbation levels, VQUNet outperforms other state-of-the-art defense methods on both CIFAR10 and Fashion-MNIST datasets with a noticeable margin. In addition, when there is no adversarial attack, only VQUNet controls its accuracy degradation $< 1\%$ from the target model's original accuracy performance compared to other methods.